%% file: name.tex
\documentclass{article}
\usepackage[utf8]{inputenc}
\usepackage[T1]{fontenc}
\usepackage{url}
\usepackage{booktabs}
\usepackage{amsfonts}
\usepackage{nicefrac}
\usepackage{microtype}
\usepackage{xcolor}
\usepackage{amsmath,amssymb,amsthm}
\usepackage{mathtools}
\usepackage{algorithm}
\usepackage{algorithmic}
\usepackage{graphicx}
\usepackage{enumitem}
\usepackage[numbers,sort&compress]{natbib}
\usepackage{rotating}
\usepackage{listings}
\usepackage[section]{placeins}
\usepackage{hyperref}
\usepackage{cleveref}
\graphicspath{{images/}}

\newtheorem{theorem}{Theorem}
\newtheorem{lemma}[theorem]{Lemma}
\newtheorem{proposition}[theorem]{Proposition}
\newtheorem{corollary}[theorem]{Corollary}
\theoremstyle{definition}

\DeclareMathOperator{\Hypergeometric}{Hypergeometric}

\DeclareMathOperator{\E}{\mathbb{E}}

\newcommand\our{Prof-K}

\title{\our{}: Probabilistic One-Pass Filtering for Efficient Top-$k$ Selection}
\author{
Tadeusz Dziarmaga\thanks{These authors contributed equally to this work.} \\
Jagiellonian University
\and
Witold Sikora\footnotemark[1] \\
Jagiellonian University
\and
\L{}ukasz Struski \\
Jagiellonian University
\and
Jacek Tabor \\
Jagiellonian University
\and
Marcin Mazur \\
Jagiellonian University
}

\date{}

\begin{document}
\maketitle

\input{sections/00_abstract}
\input{sections/01_introduction}
\input{sections/02_related_work}
\input{sections/03_method}
\input{sections/04_theory}

\input{sections/05_experiments}
\input{sections/06_conclusion}

\bibliographystyle{unsrtnat}
\bibliography{ref}

\clearpage
\appendix
\input{appendix/A_proofs}
\input{appendix/B_fast_parameters}
\input{appendix/C_implementation}
\input{appendix/D_experimental_details}
\input{appendix/E_ablations}

\end{document}

%% file: sections/00_abstract.tex
\begin{abstract}
Top-$k$ selection is a fundamental computational primitive with applications spanning databases, information retrieval, signal processing, and modern machine learning workloads, including sparse activations and attention pruning. As data sizes grow, existing approaches become inefficient: exact methods incur high memory and compute overhead, while approximate methods often rely on brittle heuristics that degrade under adversarial or heavy-tailed inputs. In this paper, we introduce Prof-K, a fast, scalable, and distribution-agnostic algorithm for exact top-$k$ selection. Prof-K performs a single-pass filtering procedure: a small random sample estimates an adaptive threshold, the $N$ input elements are streamed once into a compact buffer, and an exact top-$k$ routine on this buffer recovers the true top-$k$ elements on the first attempt with probability at least $1-\varepsilon$, where $\varepsilon>0$ is user specified. We derive high-probability guarantees for correctness and buffer size, together with an approximately optimal sample size that minimizes overhead as a function of $N$ and $k$. Empirically, Prof-K achieves $1.5\times$--$15\times$ speedups over the highly optimized PyTorch \texttt{topk} and recent RadiK implementations, with the largest gains in the large-scale, small-to-moderate-$k$ regime where prior methods struggle most. Unlike previous approaches, these guarantees hold independently of the input distribution, ensuring robustness to adversarial settings. A run-time check detects the rare failures and triggers a retry, so the returned set is always exact and $\varepsilon$ bounds only the probability of requiring an additional pass. We further demonstrate its impact on training BatchTopK Sparse Autoencoders (SAEs), where top-$k$ selection constitutes a significant portion of the training cost.
\end{abstract}

%% file: sections/01_introduction.tex
\begin{figure}[t]
\centering
\includegraphics[width=0.73\textwidth]{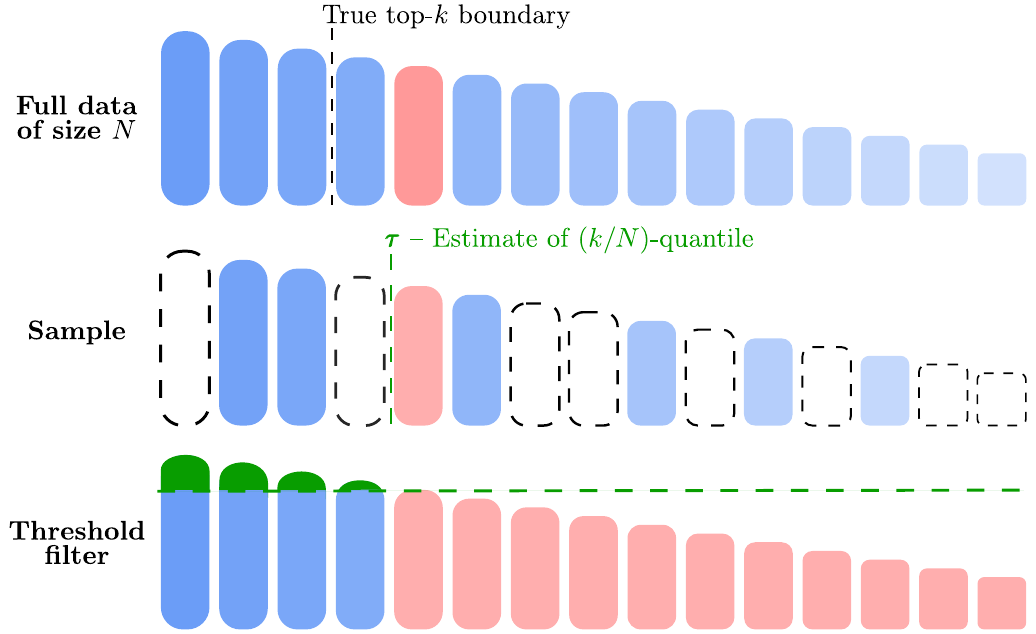}
\caption{
\textbf{The core idea of our \our{} algorithm.}
\textit{Top:} Full data of size $N$ sorted in descending order; the true top-$k$ boundary is marked by a vertical dashed line.
    \textit{Middle:} A random sample (solid outlines) is used to compute a threshold $\tau$ (green dashed line) as a conservative estimate of the $(k/N)$-quantile.
    \textit{Bottom:} The threshold filter is applied in parallel to the full data. Elements $\geq \tau$ (green) are retained in a compact candidate buffer, while elements below $\tau$ are discarded. This approach ensures all true top-$k$ elements are preserved with high probability while drastically reducing the workload for the final selection stage.}%
\label{fig:teaser}
\end{figure}

\section{Introduction}
\label{sec:introduction}

Top-$k$ selection is a core primitive of large-scale machine learning. It arises whenever only a small subset of the largest elements must be retained from a much larger collection~\cite{akbarinia2011best,shanbhag2018efficient}, for example in sparse autoencoders~\cite{bussmann2024batchtopk} and gradient sparsification~\cite{lin2018deep}. As tensor sizes continue to grow, the cost of repeated top-$k$ execution becomes increasingly significant in large-scale GPU workloads, particularly in sparse computation settings where only a small fraction of values is ultimately retained~\cite{shazeer2017sparsely,lepikhin2020gshard,fedus2022switch,bussmann2024batchtopk}.

The challenge is most visible in the large-$N$, small-$k$ regime. Even when only a tiny fraction of elements is needed, exact algorithms must still determine the precise global decision boundary across the full tensor. On modern GPUs, this cost is dominated primarily by memory movement and synchronization, making top-$k$ disproportionately expensive relative to the surrounding computation~\cite{alabi2012fast,gaihre2021dr,zhang2023parallel,xie2024rtop,li2024radik}. Radix-based kernels such as RadiK~\cite{li2024radik} reduce the latency but require auxiliary memory proportional to the input,  and cannot run at all once the input approaches the device memory. In training workloads, where the selected tensor shares device memory with weights, activations and optimizer state, this overhead directly limits the admissible model or batch size.

In this work, we introduce \our{}, a probabilistic one-pass filtering algorithm for exact top-$k$ selection on GPUs. The method draws a small uniform sample to estimate a conservative threshold, streams once through the input while retaining only the elements above this threshold in a compact candidate buffer, and applies an exact top-$k$ routine to the candidates (\Cref{fig:teaser}). The filter kernel counts the retained elements exactly, so the rare case in which the buffer is inadequate is detected at run time and handled by redrawing the sample: the output is always the exact top-$k$, and the user-specified budget $\varepsilon$ bounds the frequency of the additional pass rather than the accuracy of the result. This distinguishes \our{} from approximate top-$k$ methods, which trade accuracy for speed~\cite{chern2022tpu}. Because the analysis depends only on the ranks induced by uniform sampling, the guarantee holds for every input, including heavy-tailed and adversarial ones.

We summarize our contributions as follows.
\begin{itemize}[leftmargin=*, nosep]
    \item We propose \our{}, a sampling-based filtering algorithm for exact top-$k$ selection over the large tensors that arise in sparse ML workloads. It always returns the exact result: failures are detected at run time and resolved by resampling, with the budget $\varepsilon$ bounding only their frequency, and the guarantee holds for every input distribution. We derive the optimal parameters for a prescribed failure budget, together with closed-form approximations that are accurate and cheap to evaluate.
    \item We evaluate \our{}, implemented in Triton, against \texttt{torch.topk} and RadiK, the state-of-the-art radix selection, on a DGX H100 and an RTX 3060. \our{} is the fastest method in every large-$N$ configuration on both GPUs, by up to $16\times$ over \texttt{torch.topk} and $3.1\times$ over RadiK, and its speed is unaffected by the input distribution, including inputs adversarial for radix selection. Its auxiliary memory is measured in megabytes where RadiK requires gigabytes, which extends exact selection to inputs on which radix selection cannot run, allowing to train larger models with larger batches on the same hardware.
    \item We demonstrate the utility of \our{} in two machine learning workloads. In BatchTopK sparse autoencoders~\cite{bussmann2024batchtopk} trained on GPT-2 XL activations it reduces the end-to-end step time by $1.20$ to $1.30\times$, close to the ceiling attainable by any selection kernel; in top-$k$ gradient sparsification of GPT-2 XL it reduces the step time by $1.38\times$ with results identical to \texttt{torch.topk}. The same batch-global selection structure appears in Matryoshka and crosscoder variants of BatchTopK~\cite{bussmann2025matryoshka,minder2025overcoming}.
\end{itemize}

%% file: sections/02_related_work.tex
\section{Related Work}
\label{sec:related}

Classical exact top-$k$ methods are based on sorting, heaps, or partition-based selection such as quickselect, and efficient top-$k$ query execution has long been studied in database and parallel hardware settings~\cite{akbarinia2011best,shanbhag2018efficient}. In large-scale GPU workloads the dominant cost is memory movement rather than comparisons, which makes exact top-$k$ expensive even in optimized libraries such as PyTorch.

A substantial line of work studies efficient exact top-$k$ selection on GPUs. Early work by \citet{alabi2012fast} developed GPU-specific $k$-selection algorithms optimized for massively parallel architectures. Dr. Top-k~\cite{gaihre2021dr} introduces a delegate-centric design that improves load balancing and reduces redundant work among GPU threads. \citet{zhang2023parallel} provide a comprehensive study of parallel GPU top-$k$ algorithms and propose optimized implementations for different operating regimes. RTop-K~\cite{xie2024rtop} focuses on efficient row-wise top-$k$ selection and demonstrates strong performance for moderate-size neural network workloads. Radix-based approaches such as RadiK~\cite{li2024radik} further exploit floating-point representations to accelerate exact selection. In contrast, \our{} reduces the size of the exact selection problem itself through probabilistic filtering, making it particularly effective for very large $N$ and relatively small $k$.

Approximate and probabilistic selection methods reduce computational cost using sampling, sketching, or approximate quantile estimation, often at the cost of weaker guarantees or assumptions on the input distribution. Quantile summary methods such as the Greenwald--Khanna~\cite{greenwald2001space} algorithm provide efficient streaming approximations with bounded error guarantees. \our{} differs in that its guarantees depend only on the combinatorial properties of uniform sampling without replacement. The analysis therefore depends on ranks rather than values, yielding distribution-agnostic probabilistic guarantees.

Top-$k$ selection is increasingly important in sparse machine learning workloads, including mixture-of-experts routing, sparse attention, activation pruning, and sparse autoencoders~\cite{shazeer2017sparsely,lepikhin2020gshard,fedus2022switch,bussmann2024batchtopk,bussmann2025matryoshka,minder2025overcoming}. \our{} targets the subset of these workloads that perform
batch-global selection over very large tensors with small retained fractions. Rather than replacing exact selection entirely, \our{} uses sampling to construct a conservative threshold and then applies exact top-$k$ refinement on a compact candidate buffer, making it complementary to optimized GPU kernels such as Dr. Top-k, RTop-K, and RadiK.

%% file: sections/03_method.tex
\section{Problem Setup and Algorithm Overview}
\label{sec:method}
Given an input vector $x \in \mathbb{R}^N$ and an integer $k \geq 1$, our goal is to return the $k$ largest entries of $x$ together with their indices. For batched inputs with batch size $B$, the procedure is applied in parallel to each batch element.

\paragraph{The \our{} Algorithm}
\our{} proceeds in four stages, designed to minimize memory traffic while providing probabilistic correctness guarantees:
\begin{enumerate}
    \item \textbf{Sampling.} Uniformly sample $S$ indices without replacement from $[N] = \{1,\dots,N\}$. Let $\mathcal{S}\subset[N]$ denote the sampled indices.
    \item \textbf{Threshold Estimation.} Sort the sampled values $x|_{\mathcal S}$ in descending order. For a chosen rank $t \in [S]$, define the threshold $\tau = s_t$, where $s_t$ is the $t$-th largest sampled value.
    \item \textbf{Filtering.} Stream through the full input once, appending every element satisfying $x_i \ge \tau$ to a candidate buffer $\mathcal B$ of capacity $M$, which is usually small relative to the full input size $N$.
    \item \textbf{Refinement.} Run an exact top-$k$ routine on the buffer $\mathcal B$ to recover the final answer. If fewer than $k$ candidates are collected, or if the buffer overflows before the scan completes, \our{} redraws the sample and repeats the pass. Each attempt fails with probability at most $\varepsilon$ independently of the previous ones, so the expected number of attempts is at most $1/(1-\varepsilon)$; an implementation may also fall back to an exact top-$k$ over the full input after a fixed number of failed attempts, say six, an event of probability at most $\varepsilon^6$.
\end{enumerate}
The central design goal is to choose the parameters $(S, t, M)$ so that the number of retained elements $R = |\{i \in [N] : x_i \ge \tau\}|$ satisfies $k \le R \le M$ with probability at least $1-\varepsilon$, where $0<\varepsilon\ll 1$ is user specified. When this holds, the final exact top-$k$ operates on a small buffer rather than all $N$ entries. The idea behind the threshold is that the $(t/S)$-quantile of a uniform sample estimates the $(t/S)$-quantile of the input, so choosing $t$ slightly above $Sk/N$ yields a conservatively low threshold that all true top-$k$ elements pass with high probability while most other elements are excluded.

%% file: sections/04_theory.tex
\section{Theoretical Analysis}
\label{sec:theory}
We now formalize the probabilistic guarantees underlying \our{}. The analysis proceeds in four steps. First, we characterize the distribution of the rank that the sampled threshold occupies in the full input. Second, we prove a theorem that yields the exact optimal threshold rank $t^\star$ and buffer capacity $M^\star$ as functions of $N$, $k$, $S$ and $\varepsilon$, without any approximation. Third, we give accurate and inexpensive approximations to these two quantities for settings in which the parameters must be recomputed frequently. Fourth, we derive an asymptotically optimal choice of the sample size $S$ as a function of $N$ and $k$ that minimizes the computational overhead of the method. All proofs are given in Appendix~\ref{app:proofs}.

\paragraph{Notation and Conventions}
Throughout this section the input $x\in\mathbb R^N$ is fixed and its entries are assumed to be distinct, so that every entry has a well-defined rank, with rank $1$ assigned to the largest entry. For $K\in\{0,1,\dots,N\}$ we write $T_K\subset[N]$ for the set of indices of the $K$ largest entries of $x$. The sampled index set $\mathcal S\subset[N]$ is drawn uniformly at random among all subsets of $[N]$ of size $S$; this is what sampling $S$ indices uniformly without replacement means. As in \Cref{sec:method}, $\tau$ denotes the $t$-th largest sampled value and $R$ its rank in the full input, so that $R=|\{i\in[N]:x_i\ge\tau\}|$ is the number of elements that pass the filter.

For integers $0\le K\le N$ and $0\le S\le N$, $X\sim\Hypergeometric(N,K,S)$ is the number of distinguished items among $S$ items drawn uniformly without replacement from $N$ items of which $K$ are distinguished, that is,
\begin{equation}
\Pr(X=j)=\frac{\binom{K}{j}\binom{N-K}{S-j}}{\binom{N}{S}},
\qquad j\in\{0,1,\dots,S\},
\label{eq:hypergeom_pmf}
\end{equation}
with $\binom{a}{b}=0$ whenever $b<0$ or $b>a$. We write $F_{N,K,S}(j)=\Pr(X\le j)$ for its distribution function and
\begin{equation}
F^{-1}_{N,K,S}(p)=\min\{\,j\in\{0,1,\dots,S\}: F_{N,K,S}(j)\ge p\,\},\qquad p\in(0,1],
\label{eq:hypergeom_quantile}
\end{equation}
for its quantile function, which takes a probability and returns an integer; both are available in standard numerical libraries.

\paragraph{Sampled Counts in a Top-$K$ Set}
The single quantity on which the analysis rests is the number of sampled indices that fall into a top-$K$ set. For $K\in\{0,1,\dots,N\}$ let $X_K=|\mathcal S\cap T_K|$, and define
\begin{equation}
G(K)=\Pr\bigl(\text{fewer than $t$ of the $S$ sampled indices lie in $T_K$}\bigr)=\Pr(X_K\le t-1).
\label{eq:G_def}
\end{equation}
The function $G$ depends on $N$, $S$ and $t$ as well, and we make this dependence explicit by writing $G(N,K,S,t)$ whenever it matters.
\begin{lemma}
\label{lem:G}
For every $K\in\{0,1,\dots,N\}$, $X_K\sim\Hypergeometric(N,K,S)$ and consequently
\begin{equation}
G(K)=F_{N,K,S}(t-1).
\label{eq:G_hypergeom}
\end{equation}
Moreover, $G(K)$ is nonincreasing in $K$ and nondecreasing in $t$.
\end{lemma}

\paragraph{Failure Modes}
Exactly $R$ elements pass the filter. If $R\ge k$, the $k$-th largest entry of $x$ is at least $\tau$, so every true top-$k$ entry passes; if in addition $R\le M$, every element that passes is stored, and the refinement step returns the exact top-$k$. \our{} verifies the condition $k\le R\le M$ at run time, since the filter kernel counts the elements above the threshold exactly, and treats its violation as a failure. There are two ways in which this can happen:
\begin{enumerate}[leftmargin=*, nosep]
\item \textbf{Insufficient recall}, $A=\{R<k\}$: the threshold is too high and fewer than $k$ elements pass the filter.
\item \textbf{Buffer overflow}, $B=\{R>M\}$: more elements pass the filter than the buffer can hold, so some true top-$k$ entries may fail to enter it.
\end{enumerate}
Since $M\ge k$, the two events are disjoint and $\Pr(A\cup B)=\Pr(A)+\Pr(B)$. An overflow does not by itself make the output wrong, since the true top-$k$ entries may still be among the $M$ stored elements, so the probability of an incorrect output is bounded by $\Pr(A)+\Pr(B)$. We now compute both terms exactly.

\begin{proposition}[Insufficient recall]
\label{prop:recall}
Let $1\le k\le N$ and $1\le t\le S$. Then for every input $x$,
\begin{equation}
\Pr(R<k)=1-G(k-1)=1-F_{N,k-1,S}(t-1).
\label{eq:prob_A}
\end{equation}
\end{proposition}

\begin{proposition}[Buffer overflow]
\label{prop:overflow}
Let $1\le t\le S$ and $k\le M\le N$. Then for every input $x$,
\begin{equation}
\Pr(R>M)=G(M)=F_{N,M,S}(t-1).
\label{eq:prob_B}
\end{equation}
\end{proposition}

Neither probability depends on the values of $x$: only $N$, $k$, $S$, $t$ and $M$ enter \Cref{eq:prob_A,eq:prob_B}. This is the precise sense in which the guarantees of \our{} are distribution-free.

\paragraph{Exact Optimal Parameters}
Given a failure budget for each mode, \Cref{prop:recall,prop:overflow} reduce the choice of $t$ and $M$ to two inequalities in $G$, and the monotonicity established in \Cref{lem:G} allows both to be solved exactly. The following theorem gives the smallest threshold rank and buffer capacity under which \our{} retrieves the exact top-$k$ with a user-specified probability on the first attempt.

\begin{theorem}[Exact optimal parameters]
\label{thm:exact_params}
Let $1\le k\le N$, $1\le S\le N$ and $\varepsilon_A,\varepsilon_B\in(0,1)$ with $\varepsilon_A\ge\Pr(X_{k-1}=S)$, and set
\begin{equation}
\begin{aligned}
t^\star&=F^{-1}_{N,k-1,S}(1-\varepsilon_A)+1,\\
M^\star&=\min\bigl\{\,M\in\{k,\dots,N\}: F_{N,M,S}(t^\star-1)\le\varepsilon_B\,\bigr\}.
\end{aligned}
\label{eq:exact_params}
\end{equation}
Then $1\le t^\star\le S$, the minimum defining $M^\star$ exists, and for every input $x$:
\begin{enumerate}[leftmargin=*, nosep]
\item[(i)] $t^\star$ is the smallest threshold rank $t\in[S]$ for which the recall failure has probability $\Pr(R<k)\le\varepsilon_A$;
\item[(ii)] with $t=t^\star$, $M^\star$ is the smallest buffer capacity $M\ge k$ for which the overflow has probability $\Pr(R>M)\le\varepsilon_B$;
\item[(iii)] with $t=t^\star$ and $M=M^\star$, \our{} passes its run-time check, and hence returns the exact top-$k$, with probability at least $1-\varepsilon_A-\varepsilon_B$; the exact failure probability is
\begin{equation}
\Pr(A\cup B)=\Pr(R<k)+\Pr(R>M^\star).
\label{eq:total_failure}
\end{equation}
\end{enumerate}
\end{theorem}

The condition on $\varepsilon_A$ only excludes a degenerate case: $X_{k-1}=S$ is the event that the entire sample falls within the top $k-1$ entries, and if it had probability above $\varepsilon_A$, no threshold rank $t\le S$ could meet the recall budget. Its probability $\binom{k-1}{S}/\binom{N}{S}$ is zero for $S\ge k$ and at most $((k-1)/N)^S$ otherwise, far below any budget of interest. With the default split $\varepsilon_A=\varepsilon_B=\varepsilon/2$, \Cref{thm:exact_params} states that \our{} run with the parameters $(S,t^\star,M^\star)$ returns the exact top-$k$ on the first attempt with probability at least $1-\varepsilon$, for every input. The two parameters are also jointly optimal in the following sense.

\begin{corollary}[Joint optimality]
\label{cor:joint}
Under the assumptions of \Cref{thm:exact_params}, every pair $(t,M)$ with $t\in[S]$ and $M\in\{k,\dots,N\}$ satisfying $\Pr(R<k)\le\varepsilon_A$ and $\Pr(R>M)\le\varepsilon_B$ has $t\ge t^\star$ and $M\ge M^\star$.
\end{corollary}

Computing the pair is inexpensive: $t^\star$ is one quantile evaluation, and since $G$ is nonincreasing in $M$, $M^\star$ is found by binary search over $\{k,\dots,N\}$ with at most $\lceil\log_2 N\rceil$ evaluations of the distribution function.

\paragraph{Fast Approximate Parameter Computation}
\label{par:approx}
Evaluating \Cref{eq:exact_params} costs about a millisecond with a numerical library, comparable to the kernel itself. In most machine learning workloads the input shape and the sparsity level are fixed throughout training, so $(t^\star,M^\star)$ are computed once per configuration and cached; our implementation memoizes them. For configurations that change from call to call, Appendix~\ref{app:fast_params} derives closed-form surrogates (a skewness-corrected normal or a Poisson quantile for $t^\star$, a Wilson--Hilferty quantile for $M^\star$) that need only elementary arithmetic, are conservative in every configuration tested, and lie within a few percent of $M^\star$; \Cref{alg:params} states the complete procedure.

\begin{algorithm}[t]
\caption{Fast approximate parameter computation (budget split $\varepsilon_A=\varepsilon_B=\varepsilon/2$)}
\label{alg:params}
\begin{algorithmic}[1]
\REQUIRE $N$, $k$, failure budget $\varepsilon$; sample size bounds $S_{\min}$, $S_{\max}$; calibrated cost ratio $r(\cdot)$
\ENSURE sample size $S$, threshold rank $t$, buffer capacity $M$
\STATE $z\gets\Phi^{-1}(1-\varepsilon/2)$ \COMMENT{rational approximation}
\STATE $\kappa\gets z^2+\tfrac12$;\quad $r\gets r(k)$ \COMMENT{calibrated cost ratio}
\STATE $S_1\gets(z/r)^{2/3}(kN)^{1/3}$;\quad $S_2\gets(\kappa N/r)^{1/2}$ \COMMENT{\Cref{prop:sample_size}}
\STATE $S\gets\mathrm{clip}\bigl(\lfloor\max(S_1,S_2)\rfloor,\;S_{\min},\;S_{\max}\bigr)$ \COMMENT{$S_{\min}$ if $r=\infty$ or $k/N>2^{-10}$}
\STATE $\mu\gets(k-1)S/N$
\IF{$\mu\ge15$}
  \STATE $\sigma\gets\sqrt{\mu\,(1-\tfrac{k-1}{N})\,\tfrac{N-S}{N-1}}$;\quad $t\gets\lceil\mu+z\sigma+(z^2-1)/6\rceil+1$ \COMMENT{\Cref{eq:t_cf}}
\ELSE
  \STATE $p\gets e^{-\mu}$;\quad $c\gets p$;\quad $j\gets0$
  \WHILE{$c<1-\varepsilon/2$}
    \STATE $j\gets j+1$;\quad $p\gets p\,\mu/j$;\quad $c\gets c+p$ \COMMENT{Poisson quantile, \Cref{eq:t_poisson}}
  \ENDWHILE
  \STATE $t\gets j+1$
\ENDIF
\STATE $h\gets1/(9t)$;\quad $M\gets\bigl\lceil\tfrac{N+1}{S+1}\,t\,(1-h+z\sqrt h)^3\bigr\rceil$ \COMMENT{Wilson--Hilferty, \Cref{eq:M_wh}}
\RETURN $S,\,t,\,M$
\end{algorithmic}
\end{algorithm}

\paragraph{Choice of the Sample Size}
\label{par:sample_size}
\Cref{thm:exact_params} fixes $t^\star$ and $M^\star$ for a given $S$ but leaves $S$ free, and $S$ controls a trade-off: a larger sample makes the threshold more accurate and the buffer smaller, at the price of more sampling work. We choose $S$ by minimizing a cost model of the stages of \our{} that charges the random gather of the $S$ sampled elements and the selection of the top $t$ among them against the exact selection over the $M^\star(S)$ buffer slots; the filtering pass streams the input once whatever $S$ is. With the leading-order behavior $M^\star(S)\approx k+2z\sqrt{kN/S}+\kappa N/S$, where $z=\Phi^{-1}(1-\varepsilon/2)$ for the standard normal distribution function $\Phi$ and $\kappa=z^2+\frac12$, the model has a closed-form minimizer up to a factor below $1.6$ (\Cref{prop:sample_size} in Appendix~\ref{app:sample_size}): the optimal sample size scales as $(kN)^{1/3}$ for large $k$ and as $\sqrt N$ below, and its single hardware constant, the ratio $r$ between the per-element costs of sampling and of selecting from the buffer, is calibrated once per GPU (Appendix~\ref{app:ablation_S}). The resulting rule, clipped to $[S_{\min},S_{\max}]$ to reflect the fixed launch cost of the selection and to keep the sampling stage negligible against the scan, is given in lines~3 and~4 of \Cref{alg:params}, which collects the complete parameter computation. Since the latency is flat over a wide range of $S$ around the optimum (Appendix~\ref{app:ablation_S}), sweeping $S$ and $\varepsilon$ once for a fixed setup and caching the best values is an equally legitimate choice when the configuration does not change.

%% file: sections/05_experiments.tex
\section{Experiments and Evaluation}
\label{sec:experiments}
We compare \our{} against \texttt{torch.topk} and RadiK~\cite{li2024radik}, the state-of-the-art radix selection and the strongest exact baseline at large $N$, on a data-centre GPU (NVIDIA DGX H100, 80\,GB) and a consumer GPU (NVIDIA RTX 3060 Laptop, 6\,GB). \our{} runs at $\varepsilon=10^{-3}$ with parameters computed once per configuration and cached. The grid covers $N$ from $2^{21}$ to $2^{30}$ and $k$ from $2^{5}$ to $2^{19}$ on the H100, extended to $N=2^{31}$ against \texttt{torch.topk} alone, and $N$ up to $2^{28}$ and $k$ up to $2^{18}$ on the RTX 3060, whose memory limits the input size. Inputs are standard normal \texttt{float32} tensors; latencies are means over $30$ input tensors and $30$ timed calls each, measured with CUDA events after a warm-up, and auxiliary memory is the peak allocation above the baseline during one call. Other input distributions, including ones adversarial for radix selection, are covered in Appendix~\ref{app:ablation_dist}, further details in Appendix~\ref{app:experiments}, and ablations on batched inputs in Appendix~\ref{app:ablations}.

\begin{figure}[t]
\centering
\includegraphics[width=0.46\linewidth]{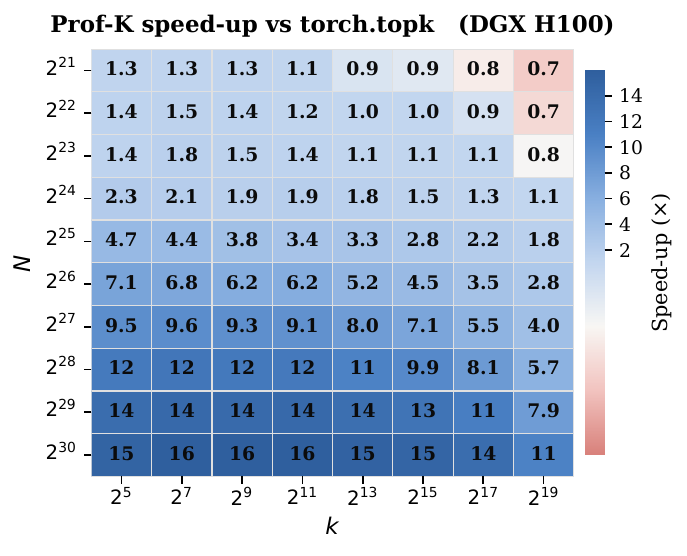}\hfill
\includegraphics[width=0.46\linewidth]{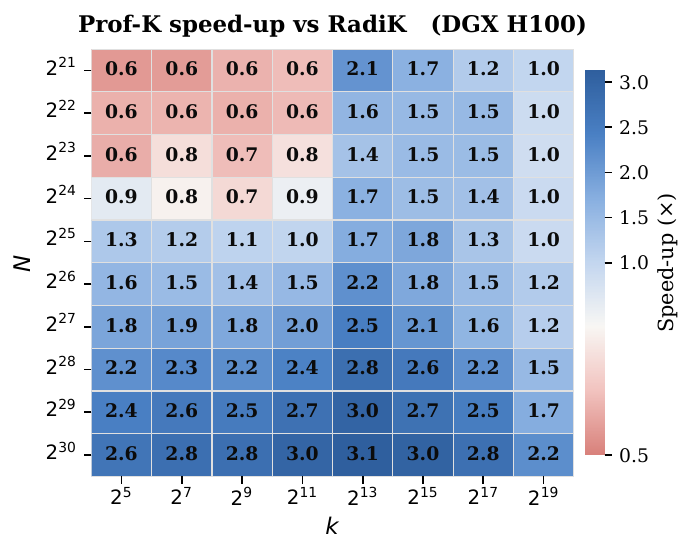}\\[1pt]
\includegraphics[width=0.46\linewidth]{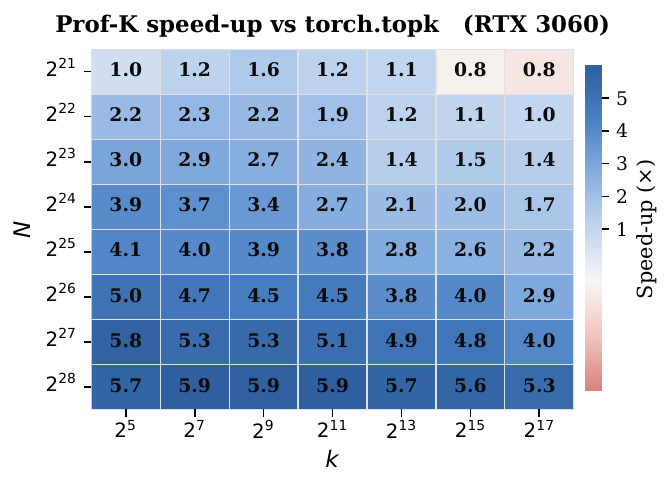}\hfill
\includegraphics[width=0.46\linewidth]{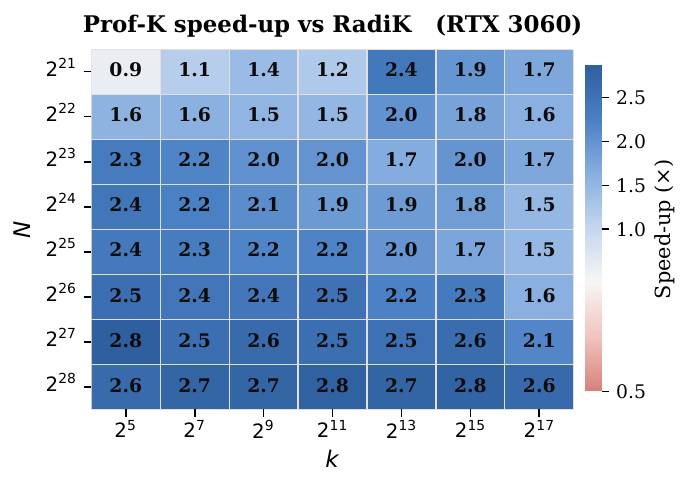}
\caption{Speed-up of \our{} over \texttt{torch.topk} (left) and RadiK (right) across the $(N,k)$ grid at $\varepsilon=10^{-3}$ on the DGX H100 (top) and the RTX 3060 (bottom). Blue marks configurations where \our{} is faster, red where it is slower.}
\label{fig:heatmaps}
\end{figure}

\begin{figure}[t]
\centering
\includegraphics[width=\linewidth]{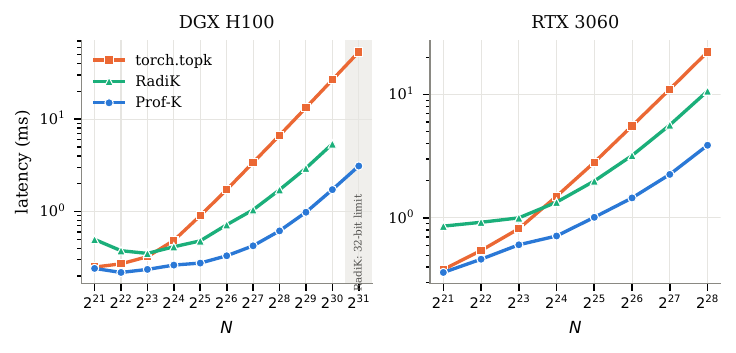}
\caption{Latency as a function of the input size at $k=2^{13}$ on both GPUs (log--log). Both baselines perform several passes over the input, \our{} a single one, so its slope is that of the memory bandwidth. On the H100 the grid extends to $N=2^{31}$ beyond the $32$-bit limit of RadiK.}
\label{fig:latency}
\end{figure}

\paragraph{Speed-up Across the $(N,k)$ Space}
\Cref{fig:heatmaps} shows the speed-up over the full grid and \Cref{fig:latency} the absolute latencies at $k=2^{13}$. Three regimes are apparent. Below $N\approx2^{23}$ every method takes a fraction of a millisecond, on the H100 between $0.2$ and $0.5$\,ms at $k=2^{13}$, and fixed costs dominate: \our{} is on par with \texttt{torch.topk} and behind RadiK for small $k$ on the H100, and ahead of both on the RTX 3060. From $N\approx2^{24}$ the streaming pass dominates every method, and the advantage of a single bandwidth-bound pass becomes visible. On the H100 at $k=2^{13}$ the latency of \our{} grows from $0.25$\,ms at $N=2^{24}$ to $1.7$\,ms at $N=2^{30}$, against $0.45$ to $26.6$\,ms for \texttt{torch.topk} and $0.42$ to $5.4$\,ms for RadiK: the speed-up over \texttt{torch.topk} rises from $1.8\times$ at $2^{24}$ to $5.2\times$ at $2^{26}$, $11\times$ at $2^{28}$ and $15\times$ at $2^{30}$, and \our{} is ahead of RadiK from $N=2^{25}$ on in every configuration but two, reaching $3.1\times$ at $2^{30}$. On the RTX 3060 the picture is the same at a smaller scale: at $N=2^{28}$ \our{} takes $3.9$\,ms against $22.1$\,ms for \texttt{torch.topk} and $10.7$\,ms for RadiK, $5.7\times$ and $2.7\times$. Finally, the dependence on $k$ is weak: at fixed $N$ the speed-up stays within a factor of about $1.3$ of its small-$k$ value up to $k\approx2^{-10}N$ and declines only beyond that, where the buffer and the refinement grow with $k$.

\begin{figure}[t]
\centering
\includegraphics[width=0.8\linewidth]{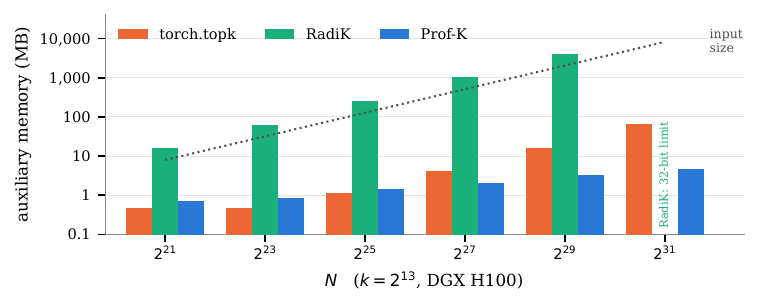}
\caption{Auxiliary GPU memory on the DGX H100 at $k=2^{13}$ (log scale); the dotted line is the size of the input. The footprints of \texttt{torch.topk} and RadiK grow with $N$, that of \our{} with the buffer capacity $M^\star$.}
\label{fig:memory}
\end{figure}

\paragraph{Memory Footprint}
\Cref{fig:memory} reports the auxiliary memory of the three methods on the H100 at $k=2^{13}$. Both baselines are $O(N)$: RadiK allocates a workspace of about $8$ bytes per input element, twice the size of the input itself, from $16$\,MB at $N=2^{21}$ to $8$\,GB at $N=2^{30}$, and \texttt{torch.topk} about $32$\,MB at $2^{30}$. The workspace of RadiK is a structural property of its radix selection, which must keep per-element state across its passes. \our{} allocates only the sample, the candidate buffer and the output, so its footprint is $O(M)$ rather than $O(N)$, and the measurements confirm it: from $0.8$\,MB at $N=2^{21}$ to $4$\,MB at $N=2^{30}$ at $k=2^{13}$, growing with $M^\star$ and not with the input, and between $2$ and $5$\,MB at $N=2^{30}$ for every $k\le2^{13}$, exceeding \texttt{torch.topk} only when $k$ approaches $2^{20}$. \our{} therefore never runs out of memory where the input itself fits, and it keeps scaling where radix selection stops: at $N=2^{31}$ it selects the top $2^{13}$ elements in $3.1$\,ms against $53.3$\,ms for \texttt{torch.topk}, a $17\times$ speed-up, and it stays above $13\times$ up to $k=2^{19}$ and above $6\times$ at $k=2^{22}$ for inputs of $3\cdot10^{9}$ and $4.3\cdot10^{9}$ elements (\Cref{tab:huge} in Appendix~\ref{app:experiments}).

\paragraph{Application: BatchTopK Sparse Autoencoders}
\label{par:sae}
To test whether kernel-level gains survive inside a training loop, we replace the batch-global top-$k$ of BatchTopK sparse autoencoders (SAEs)~\cite{bussmann2024batchtopk} with \our{}. At every step the post-encoder activations of a batch form a $B\times n$ matrix, $n$ the dictionary size, flattened into $N=B\cdot n$ entries of which the largest $B\cdot k$ are retained: one selection per step over a pool of $10^{8}$ to $10^{9}$ elements with retained fraction $k/n$, on the critical path of training.

We train BatchTopK SAEs on residual-stream activations of GPT-2 XL~\cite{radford2019language} on a DGX H100 for $10^{9}$ tokens, with dictionary sizes $n\in\{2^{16},2^{17},2^{18}\}$ and batch sizes $B$ from $4{,}096$ to $12{,}288$, so that the pool $N=B\cdot n$ ranges from $2.7\cdot10^{8}$ to $2.1\cdot10^{9}$ elements (training details in Appendix~\ref{app:experiments}). The two runs of each configuration differ only in the selection call, \texttt{torch.topk} or \our{} in exact mode, so the selected set and hence the training trajectory coincide, and the two runs are virtually identical in every tracked metric, as they should be. \Cref{tab:sae} reports the mean step time, the end-to-end speed-up and the GPU-hours of the full run. The speed-up grows with the pool size, from $1.20\times$ at $2.7\cdot10^{8}$ to $1.30\times$ at $2.1\cdot10^{9}$ elements, as the selection takes a growing share of the step, and it depends on $N$ alone: the configurations $(2^{17},8192)$ and $(2^{18},4096)$ with the same $N$ give the same gain, and the sparsity level does not enter, since top-$k$ latency is nearly flat in $k$ (\Cref{fig:heatmaps}). In absolute terms \our{} removes $0.6$ to $2.1$ GPU-hours from each $10^{9}$-token run.

\begin{table}[t]
\caption{BatchTopK SAE training on GPT-2 XL activations (DGX H100, $10^{9}$ tokens). $N=B\cdot n$ is the flattened pool size, $\rho$ is the share of the baseline step spent in top-$k$, and the last two columns give the GPU-hours of the full run.}\vskip2mm
\label{tab:sae}
\centering
\footnotesize
\setlength{\tabcolsep}{3.5pt}
\begin{tabular}{rrrrrrrrr}
\toprule
& & & \multicolumn{2}{c}{step time (ms)} & & & \multicolumn{2}{c}{GPU-hours} \\
\cmidrule(lr){4-5}\cmidrule(lr){8-9}
$n$ & $B$ & $N$ & \texttt{torch.topk} & \our{} & $\rho$ & speed-up & \texttt{torch.topk} & \our{} \\
\midrule
$2^{16}$ & $4{,}096$  & $2.7\cdot10^{8}$ & 50.3  & 41.9  & 18.5\% & $1.20\times$ & 3.42 & 2.85 \\
$2^{17}$ & $4{,}096$  & $5.4\cdot10^{8}$ & 71.4  & 58.4  & 22.4\% & $1.22\times$ & 4.84 & 3.96 \\
$2^{17}$ & $6{,}144$  & $8.1\cdot10^{8}$ & 101.8 & 81.4  & 23.6\% & $1.25\times$ & 4.60 & 3.68 \\
$2^{17}$ & $8{,}192$  & $1.1\cdot10^{9}$ & 132.2 & 104.4 & 24.2\% & $1.27\times$ & 4.48 & 3.54 \\
$2^{17}$ & $12{,}288$ & $1.6\cdot10^{9}$ & 193.0 & 150.4 & 24.9\% & $1.28\times$ & 4.36 & 3.40 \\
$2^{18}$ & $4{,}096$  & $1.1\cdot10^{9}$ & 142.8 & 115.0 & 22.4\% & $1.24\times$ & 9.69 & 7.80 \\
$2^{18}$ & $6{,}144$  & $1.6\cdot10^{9}$ & 203.6 & 161.0 & 23.6\% & $1.26\times$ & 9.21 & 7.28 \\
$2^{18}$ & $8{,}192$  & $2.1\cdot10^{9}$ & 264.4 & 203.5 & 25.1\% & $1.30\times$ & 8.97 & 6.91 \\
\bottomrule
\end{tabular}
\end{table}

\paragraph{Application: Top-$k$ Gradient Sparsification}
\label{par:gradsparse}
Top-$k$ gradient sparsification typically communicates only the largest $0.1$
to $1\%$ of the gradient at every step and keeps the rest as an error-feedback
residual~\cite{lin2018deep}. For a model with roughly $10^{9}$ parameters this is one
selection over $10^{9}$ elements per step, and its cost has long been the main
objection to the method: the selection has been measured at the same order as
the backward pass or as the communication it is meant to
save~\cite{shi2019understanding,agarwal2022utility}, and
\citet{yoon2022empirical} report uncompressed training
achieving $4.67\times$ higher throughput than exact Top-$k$ sparsification on
eight A100s, with the selection as the dominant cost.
Several widely used remedies trade exactness for speed:
thresholds estimated from a small
sample~\cite{lin2018deep} or from a Gaussian fit of the gradient
distribution~\cite{shi2019understanding}, or thresholds or
index sets periodically recomputed and reused across steps
~\cite{li2022near,zheng2025radius}. The exact primitive is still expensive:
interpolating the large-$N$ H100 measurements reported in
\Cref{tab:huge} to the $1.56\cdot10^{9}$ parameters of GPT-2 XL at
$0.1\%$ density gives estimated selection latencies of about $40$\,ms
for \texttt{torch.topk} and $4.5$\,ms for \our{}.
\our{} nevertheless returns the exact top-$k$ set at every
step.
Failures of the sampled filter are detected and handled, so the returned set remains exact.
Exact per-step selection is therefore practical in this regime
and avoids both the estimation risk of the threshold-based schemes, whose selected
count can fall far short of $k$ and stall convergence when the gradient
distribution departs from the assumed one~\cite{abdelmoniem2021efficient},
and the resampling interval of the index-reuse schemes, whose extremes degrade
the loss~\cite{zheng2025radius}.
Fine-tuning GPT-2 XL with error-feedback top-$k$
sparsification at $0.1\%$ density on a single H100 confirms this in the full
training loop; \Cref{tab:gradient-sparsification} shows a $1.38\times$
step-time speed-up for GPT-2 XL and $1.16\times$ for GPT-2 Medium, with
identical selected sets at every step.

\begin{table}[t]
  \centering
  \small
  \setlength{\tabcolsep}{5pt}
  \caption{Single-H100 fine-tuning with error-feedback Top-$k$ gradient
  sparsification at $0.1\%$ density. Both selectors return the same Top-$k$
  set at every step.}\vskip2mm
  \label{tab:gradient-sparsification}
  \begin{tabular}{lrrrr}
    \toprule
    & & \multicolumn{2}{c}{step time (ms) } & \\
    \cmidrule(lr){3-4}
    Model & Parameters & \texttt{torch.topk} & \our{} & \shortstack{speed-up} \\
    \midrule
    GPT-2 Medium & $355$M  & $85$  & $73$  & $1.16\times$ \\
    GPT-2 XL     & $1.56$B & $172$ & $125$ & $1.38\times$ \\
    \bottomrule
  \end{tabular}
\end{table}

%% file: sections/06_conclusion.tex
\section{Conclusion}
\label{sec:conclusion}
We introduced \our{}, a probabilistic one-pass filtering method for exact top-$k$ selection on GPUs. It estimates a threshold from a small sample, streams once through the input into a compact candidate buffer, and selects the top-$k$ from this buffer, while a run-time check guarantees that every output is exact. A distribution-free analysis gives the failure probability exactly and yields the smallest parameters for a prescribed budget. On a DGX H100, \our{} is up to $16\times$ faster than \texttt{torch.topk} and $3.1\times$ faster than RadiK, requires only a few megabytes of auxiliary memory, and scales to inputs that radix selection cannot process. At GPT-2 XL scale, it reduces the training time of BatchTopK sparse autoencoders and top-$k$ gradient sparsification, making it a practical building block for sparse machine learning workloads.

\paragraph{Limitations} \our{} offers no advantage on small inputs, where the fixed cost of its four stages is comparable to that of a full exact selection and RadiK remains slightly faster unless many rows are batched.  Moreover, the sample-size rule depends on a cost ratio calibrated once per GPU, and the method has not been evaluated on multi-GPU inputs, which we leave to future work.

%% file: appendix/A_proofs.tex
\section{Proofs}
\label{app:proofs}

\begin{proof}[Proof of \Cref{lem:G}]
All $\binom{N}{S}$ subsets of size $S$ are equally likely, and exactly $\binom{K}{j}\binom{N-K}{S-j}$ of them contain precisely $j$ indices of $T_K$, which gives \Cref{eq:hypergeom_pmf} for $X_K$; \Cref{eq:G_hypergeom} follows from \Cref{eq:G_def}. Monotonicity in $t$ is immediate from the definition of a cumulative distribution function. For $K<K'$ we have $T_K\subset T_{K'}$, hence $X_K\le X_{K'}$ pointwise and $\Pr(X_{K'}\le t-1)\le\Pr(X_K\le t-1)$.
\end{proof}

\begin{proof}[Proof of \Cref{prop:recall}]
The event $\{R<k\}$ is the event that $\tau$ is one of the $k-1$ largest entries of $x$. Since $\tau$ is the $t$-th largest sampled value, this happens if and only if the $t$ largest sampled values all belong to $T_{k-1}$, that is, if and only if $X_{k-1}\ge t$. Hence $\Pr(R<k)=\Pr(X_{k-1}\ge t)=1-\Pr(X_{k-1}\le t-1)=1-G(k-1)$, and \Cref{lem:G} gives the second equality.
\end{proof}

\begin{proof}[Proof of \Cref{prop:overflow}]
The event $\{R>M\}$ is the event that $\tau$ is not among the $M$ largest entries of $x$, which happens if and only if fewer than $t$ sampled indices belong to $T_M$, that is, $X_M\le t-1$. Hence $\Pr(R>M)=G(M)$, and \Cref{lem:G} gives the second equality.
\end{proof}

\begin{proof}[Proof of \Cref{thm:exact_params}]
Write $F=F_{N,k-1,S}$. By \Cref{prop:recall}, $\Pr(R<k)\le\varepsilon_A$ holds if and only if $F(t-1)\ge1-\varepsilon_A$. Since $F$ is nondecreasing, the set of such $t$ is $\{t: t-1\ge F^{-1}(1-\varepsilon_A)\}$, whose smallest element is $t^\star$; this proves (i). The quantile $F^{-1}(1-\varepsilon_A)$ is at most $S$, and it equals $S$ only if $F(S-1)<1-\varepsilon_A$, that is, $\Pr(X_{k-1}=S)>\varepsilon_A$, which the assumption excludes; hence $t^\star\le S$. For (ii), by \Cref{prop:overflow} the condition $\Pr(R>M)\le\varepsilon_B$ reads $G(M)\le\varepsilon_B$ with $t=t^\star$. By \Cref{lem:G}, $G$ is nonincreasing in $M$, so the set of $M\in\{k,\dots,N\}$ satisfying it is an interval ending at $N$; it is nonempty because $X_N=S\ge t^\star$ gives $G(N)=0$. Its smallest element is $M^\star$ by definition. Finally, $A$ and $B$ are disjoint because $M^\star\ge k$, which gives \Cref{eq:total_failure}, and the bound $\varepsilon_A+\varepsilon_B$ follows from (i) and (ii).
\end{proof}

\begin{proof}[Proof of \Cref{cor:joint}]
$t\ge t^\star$ is (i) of \Cref{thm:exact_params}. Since $G$ is nondecreasing in $t$ by \Cref{lem:G}, $G(N,M,S,t^\star)\le G(N,M,S,t)\le\varepsilon_B$, so $M$ satisfies the defining condition of $M^\star$ and $M\ge M^\star$ by minimality.
\end{proof}

\begin{proof}[Proof of \Cref{prop:sample_size}]
$\tilde C$ is a sum of a linear function and two strictly convex decreasing functions of $S$, so it is strictly convex, and $\tilde C(S)\to\infty$ as $S\to0$ or $S\to\infty$, so the minimizer exists and is unique. It is the root of $\tilde C'(S)=0$, that is, of $r=f_1(S)+f_2(S)$ with $f_1(S)=z\sqrt{kN}\,S^{-3/2}=r(S_1/S)^{3/2}$ and $f_2(S)=\kappa N S^{-2}=r(S_2/S)^{2}$, both decreasing. At $S=\max(S_1,S_2)$ one of $f_1,f_2$ equals $r$, so $f_1+f_2\ge r$ and $S^\circ\ge\max(S_1,S_2)$. At $S=2^{2/3}\max(S_1,S_2)$ we have $f_1\le r/2$ and $f_2\le 2^{-4/3}r\le r/2$, so $f_1+f_2\le r$ and $S^\circ\le2^{2/3}\max(S_1,S_2)$.
\end{proof}

\begin{proof}[Proof of \Cref{lem:ER}]
Write the sampled indices in increasing order of rank as $\rho_1<\rho_2<\dots<\rho_S$, so that $\rho_t=R$. The $S$ sampled ranks split the $N-S$ unsampled ranks into $S+1$ gaps of sizes $D_1,\dots,D_{S+1}\ge0$, where $D_i$ is the number of unsampled ranks strictly between $\rho_{i-1}$ and $\rho_i$ (with $\rho_0=0$ and $\rho_{S+1}=N+1$), and $\rho_t=t+\sum_{i\le t}D_i$. Since $\mathcal S$ is a uniformly random subset of size $S$, the vector $(D_1,\dots,D_{S+1})$ is uniformly distributed over the compositions of $N-S$ into $S+1$ nonnegative parts, a distribution that is invariant under permutations of the parts; hence the $D_i$ are exchangeable and $\E[D_i]=(N-S)/(S+1)$ for every $i$. Therefore
\[
\E[R]=t+t\,\frac{N-S}{S+1}=\frac{N+1}{S+1}\,t.
\qedhere
\]
\end{proof}

%% file: appendix/B_fast_parameters.tex
\section{Fast Parameter Computation}
\label{app:fast_params}
This appendix derives the closed-form surrogates for $t^\star$ and $M^\star$ that \Cref{alg:params} in \Cref{sec:theory} uses, derives the sample-size rule of \Cref{par:sample_size}, and compares the surrogates with the exact parameters of \Cref{thm:exact_params}. Throughout, $\varepsilon_A=\varepsilon_B=\varepsilon/2$ and $z=\Phi^{-1}(1-\varepsilon/2)$, where $\Phi$ is the standard normal distribution function; $\Phi^{-1}$ is itself evaluated by the rational approximation of \citet{acklam2003algorithm}, whose relative error is below $1.15\cdot10^{-9}$ over the whole domain, so that the fast path has no library dependence. Each surrogate is chosen according to the regime in which it is accurate, and each is designed to err on the side of a larger $t$ or a larger $M$, since an underestimate of either would violate its budget.

\textit{Threshold rank.} By \Cref{thm:exact_params}, $t^\star-1$ is the $(1-\varepsilon_A)$-quantile of $X_{k-1}\sim\Hypergeometric(N,k-1,S)$, whose mean and variance are
\begin{equation}
\mu=\frac{(k-1)S}{N},
\qquad
\sigma^2=\mu\Bigl(1-\frac{k-1}{N}\Bigr)\frac{N-S}{N-1}.
\label{eq:mu_sigma}
\end{equation}
The quantity $\mu$, the expected number of sampled indices among the top-$(k-1)$, determines the shape of the distribution and hence the appropriate approximation. When $\mu\ge15$, $X_{k-1}$ is close to normal, but it remains right-skewed, and the symmetric normal quantile $\mu+z\sigma$ systematically underestimates its upper tail. We therefore use the second-order Cornish--Fisher expansion~\cite{cornish1938moments}, which corrects the normal quantile of a distribution with skewness $\gamma$ by the term $(z^2-1)\gamma\sigma/6$. For the hypergeometric distribution $\gamma\sigma=(N-2(k-1))(N-2S)/(N(N-2))\le1$, with equality approached in the sparse regime $k\ll N$, $S\ll N$; replacing $\gamma\sigma$ by its upper bound $1$ keeps the correction conservative. This gives
\begin{equation}
t=\Bigl\lceil\mu+z\sigma+\frac{z^2-1}{6}\Bigr\rceil+1
\qquad(\mu\ge15).
\label{eq:t_cf}
\end{equation}
When $\mu<15$, the normal approximation is unreliable, but in this regime $k-1\ll N$ and $X_{k-1}$ is a count of rare events, for which $\mathrm{Poisson}(\mu)$ is the natural limit: $\Hypergeometric(N,k-1,S)$ converges to $\mathrm{Poisson}(\mu)$ as $N\to\infty$ with $(k-1)S/N=\mu$ fixed. Since $\mathrm{Poisson}(\mu)$ has variance $\mu\ge\sigma^2$, its upper quantiles dominate those of $X_{k-1}$, and replacing one by the other is again conservative. We therefore take $t=1+j^\star$ with
\begin{equation}
j^\star=\min\Bigl\{\,j\ge0:\sum_{i=0}^{j}e^{-\mu}\frac{\mu^i}{i!}\ge1-\varepsilon_A\Bigr\}
\qquad(\mu<15),
\label{eq:t_poisson}
\end{equation}
evaluated by accumulating the terms through the recurrence $p_0=e^{-\mu}$, $p_i=p_{i-1}\mu/i$, which stops after $O(\mu+z\sqrt\mu)$ iterations, that is, a few dozen at most. For $k=1$ this gives $\mu=0$ and $t=1$, which is exact: the sample maximum never exceeds the global maximum, so $\Pr(R<1)=0$.

\textit{Buffer capacity.} By \Cref{prop:overflow}, $M^\star$ is the smallest $M$ with $\Pr(R>M)\le\varepsilon_B$, that is, the $(1-\varepsilon_B)$-quantile of the rank $R$ itself. Its distribution is that of the $t$-th smallest of $S$ positions drawn uniformly without replacement from $[N]$, and its mean is known exactly.
\begin{lemma}
\label{lem:ER}
For $1\le t\le S\le N$, $\E[R]=\dfrac{N+1}{S+1}\,t$.
\end{lemma}
As $N\to\infty$ with $S$ fixed, the sampled positions become independent uniform points, and $R/(N+1)$ converges in distribution to the $t$-th order statistic of $S$ independent $\mathrm{Uniform}(0,1)$ variables; as $S\to\infty$ with $t$ fixed, this order statistic rescaled by $S+1$ converges to $\mathrm{Gamma}(t,1)$. We therefore approximate $R$ by $\frac{N+1}{S+1}\Gamma_t$ with $\Gamma_t\sim\mathrm{Gamma}(t,1)$, a scaling that matches $\E[R]$ exactly by \Cref{lem:ER}. This choice retains the feature that matters most for an upper quantile: for small $t$ the distribution of $R$ is strongly right-skewed, and a normal approximation with the same mean and variance would underestimate $M^\star$ severely, whereas $\mathrm{Gamma}(t,1)$ has the same skew. To evaluate the quantile of $\Gamma_t$ in closed form we use the Wilson--Hilferty transform~\cite{wilson1931distribution}, by which $(\Gamma_t/t)^{1/3}$ is approximately normal with mean $1-h$ and variance $h$, where $h=1/(9t)$. This yields
\begin{equation}
M=\Bigl\lceil\frac{N+1}{S+1}\,t\,\bigl(1-h+z\sqrt h\bigr)^3\Bigr\rceil,
\qquad h=\frac1{9t},
\label{eq:M_wh}
\end{equation}
for every $t$. For large $t$ the cube expands to $t+z\sqrt t+O(1)$, the normal quantile of $\mathrm{Gamma}(t,1)$, so \Cref{eq:M_wh} reduces to the normal approximation exactly where the latter is valid. At the other extreme $t=1$, where $\Gamma_1$ is exponential with $(1-\varepsilon_B)$-quantile $\ln(1/\varepsilon_B)$, \Cref{eq:M_wh} overestimates the quantile by about $3\%$ for $\varepsilon_B=5\times10^{-4}$, which is the conservative direction. The complete procedure, combining \Cref{eq:t_cf,eq:t_poisson,eq:M_wh} with the choice of $S$ derived next, is summarized in \Cref{alg:params}.

\paragraph{Choice of the Sample Size}
\label{app:sample_size}
\Cref{thm:exact_params} fixes $t^\star$ and $M^\star$ for a given $S$ but leaves $S$ free, and $S$ controls a trade-off: a larger sample makes the threshold more accurate and the buffer smaller, at the price of more sampling work. We choose $S$ by minimizing a cost model of the stages of \our{} on a GPU. The filtering pass streams the $N$ elements once at memory bandwidth and does not depend on $S$ or $M$. The other stages process a number of elements proportional to $S$ or $M$: drawing the sample is a random gather of $S$ elements at a per-element cost $g$, well above the streaming cost since each random access fetches a full memory sector; the threshold is a top-$t$ selection over the $S$ sampled values at a per-element cost $b$; and the refinement is a top-$k$ selection over the $M$ buffer slots at a per-element cost $d(k)$ that grows with $k$, since selecting many elements from the buffer is closer to a sort than to a scan. Fixed launch overheads do not affect the minimizer. The overhead beyond the mandatory scan is therefore
\begin{equation}
C(S)=(g+b)\,S+d(k)\,M^\star(S),
\label{eq:cost}
\end{equation}
where $M^\star(S)$ denotes the exact buffer capacity of \Cref{thm:exact_params} at sample size $S$. To minimize \Cref{eq:cost} in closed form we use the leading-order behavior of $M^\star(S)$, which follows from the surrogates above:
\begin{equation}
M^\star(S)\;\approx\;k+2z\sqrt{\frac{kN}{S}}+\kappa\,\frac{N}{S},
\qquad \ln(1/\varepsilon_B)\le\kappa\le z^2+\tfrac12,
\label{eq:M_asymptotic}
\end{equation}
where the lower end of the range for $\kappa$ is attained at $t=1$ and the upper end for large $t$; we use $\kappa=z^2+\frac12$. The square-root term is the familiar sampling error of a quantile estimate; the $N/S$ term is the contribution of the skew of $R$ at small $t$, and it dominates in the sparse regime $\mu\lesssim1$, where the buffer is governed by $N/S$ rather than by $k$. Substituting \Cref{eq:M_asymptotic} into \Cref{eq:cost} and writing $r=(g+b)/d(k)$ for the cost of sampling one element relative to selecting one from the buffer, the model has a unique minimizer, which we characterize as follows.

\begin{proposition}[Sample size]
\label{prop:sample_size}
Let $r,z,\kappa>0$ and let $\tilde C(S)=rS+2z\sqrt{kN/S}+\kappa N/S$ for $S>0$. Then $\tilde C$ is strictly convex with a unique minimizer $S^\circ$, and with
\begin{equation}
S_1=\Bigl(\frac{z}{r}\Bigr)^{2/3}(kN)^{1/3},
\qquad
S_2=\Bigl(\frac{\kappa N}{r}\Bigr)^{1/2},
\label{eq:S1_S2}
\end{equation}
it satisfies $\max(S_1,S_2)\le S^\circ\le2^{2/3}\max(S_1,S_2)$.
\end{proposition}

The two candidates correspond to the two terms of \Cref{eq:M_asymptotic}: for $k\gtrsim z\sqrt{rN}$ the optimal sample size scales as $(kN)^{1/3}$, and below that as $\sqrt N$, independently of $k$, because the buffer is then governed by $N/S$. We use $S=\mathrm{clip}(\max(S_1,S_2),S_{\min},S_{\max})$, which by \Cref{prop:sample_size} is within a factor $2^{2/3}\approx1.59$ of the model optimum, and the sensitivity study in Appendix~\ref{app:ablation_S} shows that the latency is flat over a considerably wider range of $S$ around it. The lower bound $S_{\min}$ reflects the fixed launch cost of the selection routine, below which a smaller sample saves nothing, and $S_{\max}$ keeps the sampling stage negligible against the scan. The only hardware-dependent quantity is the ratio $r$, which we calibrate once per GPU from a sweep over $S$ (Appendix~\ref{app:ablation_S}). The calibration confirms the two regimes the model predicts: for $k$ up to a few thousand the buffer selection costs less per element than the sampling, $r$ is large, and the rule returns the floor $S_{\min}$; for larger $k$ the buffer selection dominates, $1/r$ grows linearly in $\log_2 k$, and the sample grows with $k$ and $N$. When the retained fraction exceeds $k/N\approx2^{-10}$ the threshold selection over the sample, whose rank $t$ grows with $Sk/N$, becomes the dominant cost as well, and the floor is again the best choice.

\paragraph{Accuracy}
\Cref{tab:approx_accuracy} compares the parameters of \Cref{alg:params} with the exact optimal ones of \Cref{thm:exact_params} at $\varepsilon=10^{-3}$ over $21$ configurations spanning $N$ from $2^{20}$ to $2^{31}$ and $k$ from $2^{5}$ to $2^{20}$, with the sample size of \Cref{alg:params} in both cases. For each pair of parameters the table also reports the exact failure probability $\Pr(R<k)+\Pr(R>M)$, evaluated with the hypergeometric distribution function of \Cref{prop:recall,prop:overflow}. The surrogate threshold rank equals $t^\star$ in $19$ configurations and exceeds it by one in the other two; the surrogate buffer capacity exceeds $M^\star$ by $0.1$ to $3\%$, with the largest excess at $t=1$, where the Wilson--Hilferty quantile is least accurate, and the excess vanishes as $t$ grows. The attained failure probability is below the budget in every configuration for both parameter sets, and below that of the exact parameters for the surrogates, which confirms that the approximations err on the conservative side.

\begin{table}[!htbp]
\caption{Fast approximate parameters (\Cref{alg:params}) against the exact optimal parameters (\Cref{thm:exact_params}) at $\varepsilon=10^{-3}$, with the exact failure probability $\Pr(R<k)+\Pr(R>M)$ attained by each pair.}
\label{tab:approx_accuracy}
\centering
\footnotesize
\setlength{\tabcolsep}{3pt}
\begin{tabular}{rrr|rr|rrr|cc}
\toprule
$N$ & $k$ & $S$ & $t^\star$ & $t$ & $M^\star$ & $M$ & $M/M^\star$ & $\Pr(\text{fail})$, exact & $\Pr(\text{fail})$, approx. \\
\midrule
$2^{20}$ & $2^{5}$ & 8,192 & 4 & 4 & 1,777 & 1,801 & 1.014 & $5.99\cdot10^{-4}$ & $5.28\cdot10^{-4}$ \\
$2^{20}$ & $2^{8}$ & 8,192 & 9 & 9 & 2,835 & 2,855 & 1.007 & $7.09\cdot10^{-4}$ & $6.60\cdot10^{-4}$ \\
$2^{20}$ & $2^{11}$ & 8,192 & 32 & 32 & 6,883 & 6,909 & 1.004 & $7.53\cdot10^{-4}$ & $7.09\cdot10^{-4}$ \\
$2^{20}$ & $2^{14}$ & 8,192 & 167 & 168 & 27,141 & 27,385 & 1.009 & $9.70\cdot10^{-4}$ & $7.64\cdot10^{-4}$ \\
$2^{24}$ & $2^{5}$ & 16,384 & 2 & 2 & 10,232 & 10,422 & 1.019 & $9.35\cdot10^{-4}$ & $8.57\cdot10^{-4}$ \\
$2^{24}$ & $2^{8}$ & 16,384 & 4 & 4 & 14,259 & 14,402 & 1.010 & $6.28\cdot10^{-4}$ & $5.75\cdot10^{-4}$ \\
$2^{24}$ & $2^{11}$ & 20,650 & 10 & 10 & 19,281 & 19,362 & 1.004 & $7.91\cdot10^{-4}$ & $7.60\cdot10^{-4}$ \\
$2^{24}$ & $2^{14}$ & 36,403 & 58 & 58 & 39,777 & 39,822 & 1.001 & $8.27\cdot10^{-4}$ & $8.09\cdot10^{-4}$ \\
$2^{24}$ & $2^{17}$ & 16,384 & 168 & 168 & 218,748 & 219,092 & 1.002 & $8.87\cdot10^{-4}$ & $8.49\cdot10^{-4}$ \\
$2^{28}$ & $2^{5}$ & 16,384 & 2 & 2 & 163,769 & 166,737 & 1.018 & $5.02\cdot10^{-4}$ & $4.26\cdot10^{-4}$ \\
$2^{28}$ & $2^{8}$ & 37,480 & 3 & 3 & 86,297 & 87,385 & 1.013 & $5.07\cdot10^{-4}$ & $4.47\cdot10^{-4}$ \\
$2^{28}$ & $2^{11}$ & 82,603 & 5 & 5 & 51,044 & 51,435 & 1.008 & $9.89\cdot10^{-4}$ & $9.45\cdot10^{-4}$ \\
$2^{28}$ & $2^{14}$ & 110,643 & 18 & 18 & 85,613 & 85,777 & 1.002 & $7.37\cdot10^{-4}$ & $7.19\cdot10^{-4}$ \\
$2^{28}$ & $2^{17}$ & 234,231 & 152 & 153 & 224,429 & 225,793 & 1.006 & $9.45\cdot10^{-4}$ & $8.20\cdot10^{-4}$ \\
$2^{28}$ & $2^{20}$ & 16,384 & 93 & 93 & 2,095,476 & 2,098,101 & 1.001 & $8.83\cdot10^{-4}$ & $8.60\cdot10^{-4}$ \\
$2^{31}$ & $2^{5}$ & 16,384 & 1 & 1 & 996,031 & 1,026,231 & 1.030 & $7.36\cdot10^{-4}$ & $6.34\cdot10^{-4}$ \\
$2^{31}$ & $2^{8}$ & 106,011 & 2 & 2 & 202,532 & 206,164 & 1.018 & $5.78\cdot10^{-4}$ & $5.03\cdot10^{-4}$ \\
$2^{31}$ & $2^{11}$ & 233,638 & 4 & 4 & 128,067 & 129,279 & 1.009 & $5.86\cdot10^{-4}$ & $5.35\cdot10^{-4}$ \\
$2^{31}$ & $2^{14}$ & 262,144 & 9 & 9 & 181,989 & 182,719 & 1.004 & $7.37\cdot10^{-4}$ & $7.08\cdot10^{-4}$ \\
$2^{31}$ & $2^{17}$ & 262,144 & 32 & 32 & 441,824 & 442,225 & 1.001 & $7.76\cdot10^{-4}$ & $7.65\cdot10^{-4}$ \\
$2^{31}$ & $2^{20}$ & 262,144 & 168 & 168 & 1,752,499 & 1,752,830 & 1.000 & $9.09\cdot10^{-4}$ & $9.04\cdot10^{-4}$ \\
\bottomrule
\end{tabular}
\end{table}

%% file: appendix/C_implementation.tex
\section{Implementation Details}
\label{app:implementation}

\paragraph{The Filter Kernel}
The only custom kernel in \our{} is the filtering pass, shown in \Cref{lst:kernel}; sampling, the threshold selection over the sample and the final selection over the buffer are standard library calls (\texttt{torch.topk} on the $S$ sampled values and on the $M$ buffer slots). The kernel is written in Triton and launched on a two-dimensional grid, one axis over blocks of \texttt{BLOCK\_SIZE} consecutive elements of a row and one over the rows of the batch, so that batched inputs are handled by the same code with a per-row threshold, buffer and counter. Each program loads its block once, compares it against the threshold of its row, and counts the elements that pass. If any do, it reserves a contiguous range of buffer slots for them with a single atomic addition on the per-row counter, computes the position of each retained element within that range by an in-block prefix sum, and writes values and indices to the reserved slots. The pass therefore streams the input exactly once, issues one atomic per block rather than one per retained element, and writes only the retained elements, which is what keeps it at the memory bandwidth of the device.

The counter is incremented by the full number of retained elements in the block, whereas the stores are masked to the buffer capacity. After the pass the counter of each row therefore holds the exact rank $R$ of its threshold, whether or not the buffer overflowed, and the run-time check $k\le R\le M$ of \Cref{sec:method} is a comparison of $B$ integers, without any extra pass over the input. The block size and the number of warps are selected by Triton's autotuner from the configurations listed, keyed on the row length, and the counters are reset to zero between autotuning trials so that the trials do not accumulate into them.

\begin{figure}[p]
\begin{lstlisting}[language=Python,caption={The Triton filter kernel of \our{}.},label={lst:kernel}]
import triton
import triton.language as tl

@triton.autotune(
    configs=[
        triton.Config({'BLOCK_SIZE': 1024}, num_warps=8, num_stages=3),
        triton.Config({'BLOCK_SIZE': 2048}, num_warps=8, num_stages=4),
        triton.Config({'BLOCK_SIZE': 4096}, num_warps=8, num_stages=4),
    ],
    key=['n_elements'],
    reset_to_zero=['count_ptr'],
)
@triton.jit
def _filter_topk_batched_kernel(
    x_ptr, out_vals_ptr, out_idx_ptr, count_ptr, thresholds_ptr,
    stride_xb: tl.int64, stride_ob: tl.int64, n_elements: tl.int64, buffer_cap,
    BLOCK_SIZE: tl.constexpr,
):
    pid_chunk = tl.program_id(axis=0)   # block of the input row
    pid_batch = tl.program_id(axis=1)   # row of the batch

    # 64-bit offsets: inputs may exceed 2^31 elements
    block_start = pid_chunk.to(tl.int64) * BLOCK_SIZE
    offsets = block_start + tl.arange(0, BLOCK_SIZE)
    read_mask = offsets < n_elements

    x_offset = pid_batch.to(tl.int64) * stride_xb + offsets
    x = tl.load(x_ptr + x_offset, mask=read_mask, other=-float('inf'))
    thresh = tl.load(thresholds_ptr + pid_batch)

    pass_mask = (x >= thresh) & read_mask
    pass_int = pass_mask.to(tl.int32)
    n_pass = tl.sum(pass_int)

    if n_pass > 0:
        # one atomic per block reserves n_pass slots and counts every
        # retained element, also beyond the buffer capacity
        global_offset = tl.atomic_add(count_ptr + pid_batch, n_pass)
        local_offset = tl.cumsum(pass_int) - 1
        write_idx = global_offset + local_offset
        write_mask = pass_mask & (write_idx < buffer_cap)

        out_offset = pid_batch * stride_ob + write_idx
        tl.store(out_vals_ptr + out_offset, x, mask=write_mask)
        tl.store(out_idx_ptr + out_offset, offsets.to(tl.int64),
                 mask=write_mask)
\end{lstlisting}
\end{figure}

%% file: appendix/D_experimental_details.tex
\section{Experimental Details and Additional Results}
\label{app:experiments}

\begin{sidewaysfigure}
\centering
\includegraphics[width=0.82\linewidth]{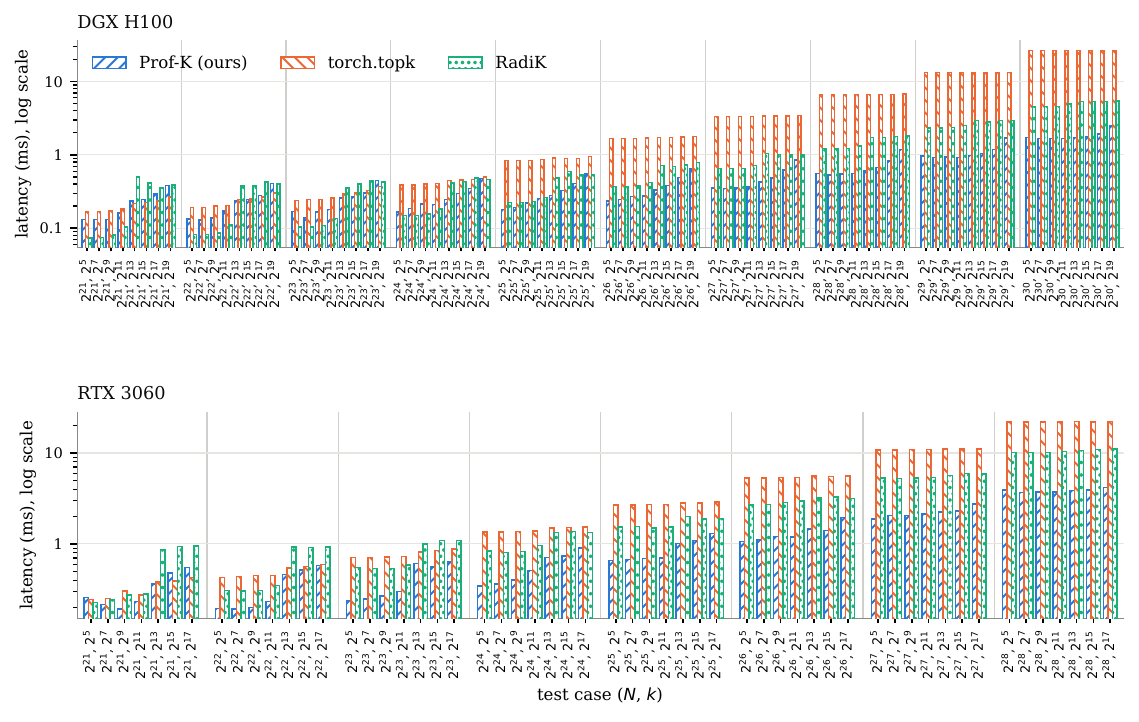}
\caption{Latency (ms, log scale) of \texttt{torch.topk}, RadiK and \our{} on every $(N,k)$ configuration of the grid, on the DGX H100 (top, $k\in\{2^{5},2^{7},\dots,2^{19}\}$) and the RTX 3060 (bottom, $k\in\{2^{5},2^{7},\dots,2^{17}\}$), at $\varepsilon=10^{-3}$. Groups are ordered by $N$ and, within a group, by $k$.}
\label{fig:latency_all}
\end{sidewaysfigure}

\paragraph{Latency of Every Configuration}
\Cref{fig:latency_all} reports the absolute latency of the three methods on every $(N,k)$ configuration of the grid on both GPUs, grouped by $N$ with $k$ increasing within each group.

\begin{figure}[t]
\centering
\includegraphics[width=\linewidth]{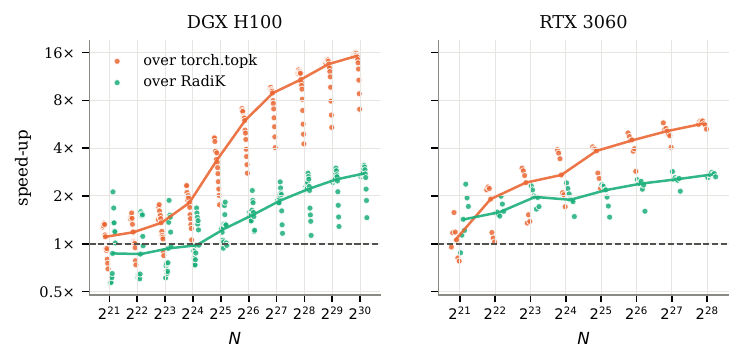}
\caption{Speed-up of \our{} as a function of $N$; each point is one $(N,k)$ configuration, horizontally spread by $k$, and the lines join the medians over $k$. The dashed line marks parity.}
\label{fig:scatter}
\end{figure}

\paragraph{Speed-up as a Function of $N$}
\Cref{fig:scatter} shows every $(N,k)$ configuration of the grid of \Cref{sec:experiments} as one point, with the spread over $k$ made visible. \Cref{tab:huge} lists the configurations beyond $N=2^{30}$ on the H100, where RadiK cannot run because its $32$-bit indexing stops at $2^{31}$ elements, so \our{} is compared with \texttt{torch.topk} alone.

\begin{table}[t]
\caption{\our{} against \texttt{torch.topk} on the DGX H100 for inputs beyond the reach of RadiK. Latency in ms, auxiliary memory in MB.}
\label{tab:huge}
\centering
\small
\setlength{\tabcolsep}{4pt}
\begin{tabular}{rrrrrrr}
\toprule
$N$ & $k$ & \shortstack{\our{}\\(ms)} & \shortstack{\texttt{torch.topk}\\(ms)} & speed-up & \shortstack{\our{}\\(MB)} & \shortstack{\texttt{torch.topk}\\(MB)} \\
\midrule
$2^{31}$ & $2^{10}$ & 3.14 & 53.8 & $17.1\times$ & 3.1 & 65.0 \\
$2^{31}$ & $2^{14}$ & 3.19 & 53.5 & $16.8\times$ & 5.8 & 65.2 \\
$2^{31}$ & $2^{18}$ & 3.56 & 53.4 & $15.0\times$ & 24.1 & 68.0 \\
$2^{31}$ & $2^{22}$ & 7.25 & 54.0 & $7.5\times$ & 238 & 177 \\
$3\cdot10^{9}$ & $2^{10}$ & 4.67 & 74.6 & $16.0\times$ & 5.4 & 91.4 \\
$3\cdot10^{9}$ & $2^{18}$ & 5.61 & 74.6 & $13.3\times$ & 32.4 & 94.4 \\
$4.3\cdot10^{9}$ & $2^{10}$ & 6.61 & 106.7 & $16.1\times$ & 5.7 & 130.0 \\
$4.3\cdot10^{9}$ & $2^{18}$ & 7.51 & 106.8 & $14.2\times$ & 35.1 & 133.0 \\
$4.3\cdot10^{9}$ & $2^{22}$ & 15.4 & 106.5 & $6.9\times$ & 285 & 178 \\
\bottomrule
\end{tabular}
\end{table}

\paragraph{Cost of the Run-Time Check}
Exact mode is guaranteed to return the true top-$k$; the price is the host synchronisation needed to inspect the count $R$ after the filtering pass, and the rare redraw. On the H100 this costs between $5\%$ of the latency at $N=2^{30}$ and $40\%$ at $N=2^{21}$, where the synchronisation exposes kernel-launch latency that is otherwise hidden. Omitting the check, which is appropriate when an occasional miss is acceptable, recovers this overhead, and we study the resulting accuracy--speed trade-off in Appendix~\ref{app:ablations}.

%% file: appendix/E_ablations.tex
\section{Ablations}
\label{app:ablations}

\subsection{Sample Size}
\label{app:ablation_S}
The sample size $S$ is the one parameter of \our{} that the exact analysis of \Cref{sec:theory} leaves free, and \Cref{prop:sample_size} chooses it from a cost model whose single hardware constant, the cost ratio $r$, has to be calibrated. We therefore sweep $S$ on the RTX 3060 to calibrate $r$ and to measure how far the resulting rule lands from the latency-optimal sample size. The sweep covers $S\in\{2^{12},2^{13},\dots,2^{18}\}$, capped at $N/32$, for every $N\in\{2^{21},\dots,2^{28}\}$ and $k\in\{2^{5},2^{8},2^{11},2^{14},2^{17}\}$; for each $S$ the threshold rank and buffer capacity are the exact optimal ones of \Cref{thm:exact_params} at $\varepsilon=10^{-3}$, and the latency is that of exact mode, averaged over $8$ input tensors and $20$ timed calls each after a warm-up.

\paragraph{Calibration}
Fitting the cost model $C=c+(g+b)S+d(k)M^\star(S)$ to the $40\times5$ measurements with $S\ge2^{14}$, with one intercept $c$ per configuration, one sampling cost $g+b$, and one buffer cost $d(k)$ per value of $k$, explains the latencies to within $4.6\%$ on average ($R^2=0.998$). The sampling cost is $0.36$\,ns per element. The buffer cost is not resolvable from zero for $k\le2^{11}$, and equals $1.6$\,ns per element at $k=2^{14}$ and $2.4$\,ns at $k=2^{17}$: selecting many elements from the buffer costs several times more per element than sampling, which is why the sample should grow with $k$. The ratio is well described by $1/r(k)=0.59\log_2 k-4.3$ for $k>2^{7}$ and $r=\infty$ below, and this is the function $r(\cdot)$ used in \Cref{alg:params}.

\begin{figure}[t]
\centering
\includegraphics[width=\linewidth]{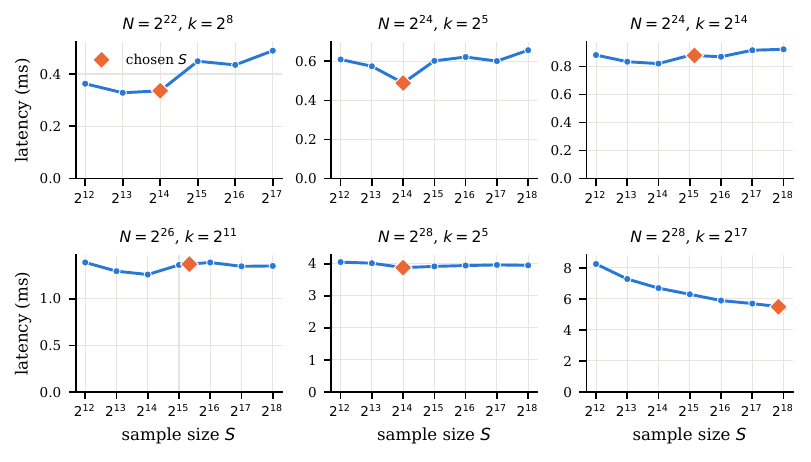}
\caption{Exact-mode latency of \our{} as a function of the sample size $S$ on the RTX 3060 for six $(N,k)$ configurations, with the sample size chosen by \Cref{alg:params} marked. The curves are flat over most of the range; at large $k$ and large $N$ a larger sample pays for itself through a smaller buffer, and the rule follows it.}
\label{fig:ablation_S}
\end{figure}

\paragraph{Sensitivity and Accuracy of the Rule}
\Cref{fig:ablation_S} shows six representative configurations and \Cref{tab:ablation_S} summarizes all $40$. First, the latency is insensitive to $S$ over a wide range: across the $64\times$ span of the sweep, the ratio between the slowest and the fastest sample size has a median of $1.24$ and never exceeds $1.51$, so a factor of two in either direction around the optimum typically costs a few percent. Second, the empirical optimum is $S=2^{14}$ in $27$ of the $40$ configurations, which is the fixed-cost floor $S_{\min}$; the exceptions are the large-$k$ configurations, where the optimum moves to $2^{18}$ as $N$ grows. Third, the sample size returned by \Cref{alg:params} lands on average $2.2\%$ above the optimal latency of the sweep, with a worst case of $15\%$ for moderate $k$ at large $N$, where the rule leaves the floor slightly earlier than the measurements warrant; run-to-run variation between sessions on this GPU is of the same order, so we do not resolve the rule further.

\begin{table}[t]
\caption{Sample-size ablation on the RTX 3060 (exact mode, $\varepsilon=10^{-3}$). For each $N$, mean over the five values of $k$ of the ratio between the slowest and the fastest sample size in the sweep, and of the latency at the sample size chosen by \Cref{alg:params} relative to the empirical optimum.}
\label{tab:ablation_S}
\centering
\small
\begin{tabular}{lcccccccc}
\toprule
$N$ & $2^{21}$ & $2^{22}$ & $2^{23}$ & $2^{24}$ & $2^{25}$ & $2^{26}$ & $2^{27}$ & $2^{28}$ \\
\midrule
slowest / fastest $S$ & 1.33 & 1.40 & 1.34 & 1.20 & 1.11 & 1.13 & 1.20 & 1.29 \\
\Cref{alg:params} / optimum & 1.01 & 1.00 & 1.01 & 1.02 & 1.04 & 1.04 & 1.05 & 1.01 \\
\bottomrule
\end{tabular}
\end{table}

\subsection{Batched Inputs}
\label{app:ablation_batched}
The main evaluation concerns a single large vector. Here we ask whether \our{} also pays off on batched inputs, where each of $B$ rows of length $N$ is selected independently in one call, and what relaxing the failure budget buys. We measure \texttt{torch.topk}, RadiK and \our{} in approximate mode on the RTX 3060 for $N\in\{2^{16},2^{18},2^{20},2^{22}\}$, $B\in\{1,16,128,1024\}$ with $B\cdot N\le2^{28}$, and $k\in\{64,1024\}$, on standard normal inputs of shape $(B,N)$, averaging over $5$ input tensors and $20$ timed calls each after a warm-up. The budget $\varepsilon$ is per call and is split as $\varepsilon/B$ across the rows, so that by the union bound the probability that any row misses is at most $\varepsilon$; the recall, the fraction of the true top-$k$ indices returned, is measured against \texttt{torch.topk}.

\begin{figure}[t]
\centering
\includegraphics[width=\linewidth]{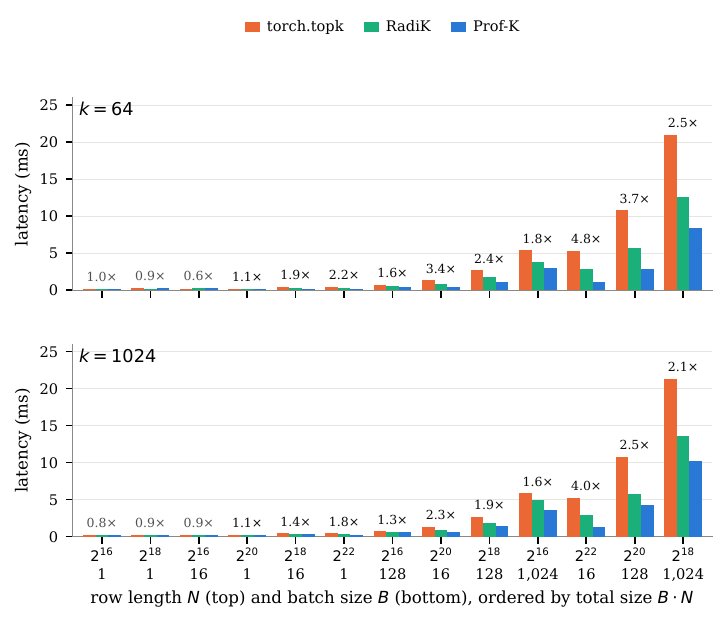}
\caption{Latency of \texttt{torch.topk}, RadiK and \our{} (approximate mode, $\varepsilon=10^{-3}$) on batched inputs on the RTX 3060, for $k=64$ (top) and $k=1024$ (bottom). Each group is one $(N,B)$ configuration, ordered by the total number of elements $B\cdot N$; the number above a group is the speed-up of \our{} over \texttt{torch.topk}.}
\label{fig:batched_latency}
\end{figure}

\paragraph{Batching Extends the Gains to Small $N$}
\Cref{fig:batched_latency} shows the latencies at $\varepsilon=10^{-3}$. For a single row of $N\le2^{20}$ no method takes more than $0.5$\,ms and the three are within $10$ to $40\%$ of each other, since the fixed costs of the four stages of \our{} are comparable to an exact selection over a million elements. Batching amortizes these fixed costs, and the total size $B\cdot N$ is what governs the outcome. From $B\cdot N=2^{23}$ on, \our{} is ahead of both baselines in every configuration, and its advantage grows with the total size: $3.4\times$ over \texttt{torch.topk} and $2.1\times$ over RadiK at $N=2^{20}$, $B=16$; $4.8\times$ and $2.6\times$ at $N=2^{22}$, $B=16$; $3.7\times$ and $2.0\times$ at $N=2^{20}$, $B=128$; and $2.5\times$ and $1.5\times$ at $N=2^{18}$, $B=1024$, all for $k=64$, with $k=1024$ between $10$ and $30\%$ lower. At a fixed total size, longer rows are more favourable than more rows, since the per-row costs of the threshold selection and of the buffer are then spread over more elements: $B\cdot N=2^{26}$ gives $4.8\times$ for $N=2^{22}$, $B=16$ but $1.8\times$ for $N=2^{16}$, $B=1024$. The crossover at $B\cdot N\approx2^{22}$ coincides with the one for a single vector in \Cref{fig:heatmaps} once the horizontal axis there is read as the total size. The recall is $1.000$ in every configuration.

\paragraph{Relaxing the Budget}
\Cref{tab:eps_recall} reports the same measurements at four budgets. Loosening $\varepsilon$ from $10^{-3}$ to $0.3$ shortens the latency by only about $5\%$ at the median, because $S$, $t^\star$ and $M^\star$ depend on $\varepsilon$ only through $z=\Phi^{-1}(1-\varepsilon/2)$, whereas the recall of single rows drops to between $0.87$ and $0.97$. On batches the split budget keeps the per-row budget small even at $\varepsilon=0.3$, and the recall stays above $0.996$. \our{} therefore has no useful accuracy knob below the exact regime, and there is no reason to run it with a loose budget: the recommended operating point is $\varepsilon=10^{-3}$, at which the approximate mode returns the exact top-$k$ in all but one call in a thousand.

\begin{table}[!htbp]
\caption{Approximate mode of \our{} on the RTX 3060 as a function of the failure budget $\varepsilon$, over the $26$ configurations of \Cref{fig:batched_latency}: latency relative to $\varepsilon=10^{-3}$ (median), and minimum and median recall for single rows ($B=1$) and for batches ($B\ge16$).}
\label{tab:eps_recall}
\centering
\footnotesize
\setlength{\tabcolsep}{3.5pt}
\begin{tabular}{lcccc}
\toprule
$\varepsilon$ & $10^{-3}$ & $10^{-2}$ & $0.1$ & $0.3$ \\
\midrule
latency relative to $\varepsilon=10^{-3}$ & 1.00 & 0.95 & 0.93 & 0.95 \\
recall, $B=1$ (min / median) & 1.000 / 1.000 & 1.000 / 1.000 & 0.966 / 0.995 & 0.871 / 0.945 \\
recall, $B\ge16$ (min / median) & 1.000 / 1.000 & 1.000 / 1.000 & 0.998 / 1.000 & 0.997 / 1.000 \\
\bottomrule
\end{tabular}
\end{table}

\subsection{Adversarial Input Distributions}
\label{app:ablation_dist}
The guarantee of \Cref{sec:theory} depends only on the ranks induced by uniform sampling, never on the values, so the latency of \our{} should be the same for every input distribution. Radix selection is not distribution-free: it discards candidates one digit of the binary representation at a time, and inputs whose values share their leading bits defeat the reduction. \citet{li2024radik} name values concentrated in a narrow range, such as uniform on $[128.6,128.7]$, as an adversarial case for RadiK and counter it with an adaptive scaling step that subtracts a randomly drawn element before the selection. We measure all methods on the RTX 3060 at $N=2^{27}$ and $k\in\{2^{8},2^{13}\}$ on seven input distributions: this adversarial case, uniform on $[1,2)$, whose values share their sign and exponent bits, and five benign ones, including the heavy-tailed log-normal and Cauchy. RadiK is run without its scaling step, as in the rest of the paper, and with it; \our{} runs in exact mode at $\varepsilon=10^{-3}$, and every returned value is verified against \texttt{torch.topk}.

\begin{table}[!htbp]
\caption{Latency (ms) on the RTX 3060 at $N=2^{27}$ as a function of the input distribution: \texttt{torch.topk}, RadiK without and with ($\dagger$) its adaptive scaling step, and \our{} in exact mode at $\varepsilon=10^{-3}$. Mean over $5$ input tensors and $20$ timed calls each.}
\label{tab:dist}
\centering
\small
\setlength{\tabcolsep}{2pt}
\begin{tabular}{l|cccc|cccc}
\toprule
& \multicolumn{4}{c|}{$k=2^{8}$} & \multicolumn{4}{c}{$k=2^{13}$} \\
input distribution & \texttt{torch} & RadiK & RadiK$^{\dagger}$ & \our{} & \texttt{torch} & RadiK & RadiK$^{\dagger}$ & \our{} \\
\midrule
standard normal & 11.2 & 5.3 & 5.3 & 2.3 & 11.5 & 5.8 & 5.8 & 2.4 \\
uniform $[0,1)$ & 11.6 & 6.1 & 5.6 & 2.2 & 11.3 & 6.6 & 6.1 & 2.4 \\
uniform $[1,2)$ & 11.2 & 6.9 & 5.7 & 2.2 & 11.4 & 7.3 & 6.2 & 2.3 \\
uniform $[128.6,128.7]$ & 10.8 & 15.8 & 5.6 & 2.2 & 11.2 & 16.4 & 6.0 & 2.4 \\
log-normal & 11.3 & 5.3 & 5.3 & 2.3 & 11.6 & 5.8 & 5.8 & 2.4 \\
exponential & 11.7 & 5.4 & 5.3 & 2.2 & 11.4 & 6.0 & 5.8 & 2.5 \\
Cauchy & 11.4 & 5.4 & 5.4 & 2.3 & 11.7 & 5.8 & 5.8 & 2.4 \\
\bottomrule
\end{tabular}
\end{table}

\Cref{tab:dist} confirms the picture. \our{} takes $2.2$ to $2.5$\,ms on every distribution, with at most $0.02$ redraws per call and exact values on every call. RadiK without scaling slows down as soon as values share their leading bits, by $15$ to $30\%$ on uniform inputs and by $3\times$ on the narrow-range input, where it falls behind \texttt{torch.topk}; the speed-up of \our{} over it grows accordingly, from $2.3\times$ on normal inputs to $7\times$ on the adversarial one. Its scaling step removes the worst case, after which \our{} is still $2.3$ to $2.6\times$ faster on every distribution. Heavy tails cost no method anything, since the top-$k$ elements then occupy their own leading digits. \our{} is thus impervious to the input distribution in practice as well as in theory, and it needs no input-dependent preprocessing to be so.

\subsection{Failure Budget}
\label{app:ablation_eps}
The failure budget $\varepsilon$ is the one user-facing parameter of \our{}. In exact mode it trades the cost of a single attempt, which grows slowly as $\varepsilon$ shrinks through $z=\Phi^{-1}(1-\varepsilon/2)$ in the sample size and the buffer capacity, against the frequency of redraws, which grows linearly in $\varepsilon$. We measure this trade-off on the RTX 3060 in exact mode with the exact parameters of \Cref{thm:exact_params}, for $\varepsilon$ from $0.3$ down to $10^{-4}$ at $N\in\{2^{24},2^{28}\}$ and $k\in\{2^{8},2^{14}\}$, recording for every call the count $R$ of the first attempt, the number of redraws and the latency of the whole call including the check. Between $2{,}000$ and $10{,}000$ calls are made per configuration, $140{,}000$ in total.

\paragraph{The Failure Probabilities Are Exact}
The count $R$ of the first attempt tells which failure mode occurred, so the experiment measures $\Pr(R<k)$ and $\Pr(R>M)$ directly. \Cref{tab:eps_failures} compares the observed numbers of recall failures and buffer overflows, pooled over the four $(N,k)$ configurations, with the numbers predicted by \Cref{prop:recall,prop:overflow} at the same parameters. The two agree at every budget within the sampling noise of the counts, from $1{,}194$ overflows observed against $1{,}200$ predicted at $\varepsilon=0.3$ to $15$ against $20$ at $10^{-3}$, and the observed total never exceeds $\varepsilon$ times the number of calls. The recall failures are rarer than $\varepsilon_A$ allows at small $k$, where the integer threshold rank cannot exhaust its budget: at $N=2^{28}$, $k=2^{8}$ and $\varepsilon=0.3$ the rank is $t^\star=1$ and $\Pr(R<k)$ is $0.015$ rather than $0.15$.

\begin{table}[!htbp]
\caption{Predicted and observed numbers of recall failures ($R<k$) and buffer overflows ($R>M$) on the first attempt, pooled over the four $(N,k)$ configurations of the failure-budget ablation (RTX 3060, exact mode). Predictions are the exact probabilities of \Cref{prop:recall,prop:overflow} multiplied by the number of calls.}
\label{tab:eps_failures}
\centering
\small
\begin{tabular}{lrrrrr}
\toprule
$\varepsilon$ & calls & \shortstack{predicted\\$R<k$} & \shortstack{observed\\$R<k$} & \shortstack{predicted\\$R>M$} & \shortstack{observed\\$R>M$} \\
\midrule
$0.3$ & 8,000 & 545.8 & 518 & 1,199.9 & 1,194 \\
$0.1$ & 8,000 & 248.0 & 245 & 399.9 & 415 \\
$0.03$ & 8,000 & 41.5 & 34 & 120.0 & 116 \\
$0.01$ & 12,000 & 31.4 & 38 & 60.0 & 51 \\
$0.003$ & 24,000 & 16.5 & 25 & 36.0 & 30 \\
$10^{-3}$ & 40,000 & 7.0 & 7 & 20.0 & 15 \\
$10^{-4}$ & 40,000 & 1.0 & 0 & 2.0 & 2 \\
\bottomrule
\end{tabular}
\end{table}

\begin{figure}[!htbp]
\centering
\includegraphics[width=\linewidth]{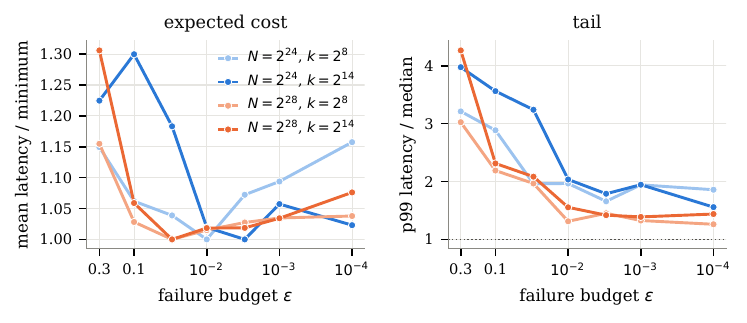}
\caption{Exact-mode latency of \our{} on the RTX 3060 as a function of the failure budget $\varepsilon$, per call including the run-time check and any redraw. Left: mean latency relative to its minimum over $\varepsilon$. Right: the $99$th percentile relative to the median.}
\label{fig:eps_ablation}
\end{figure}

\paragraph{The Expected Cost Is Flat, but the Tail Events Are Not}
\Cref{fig:eps_ablation} shows the latency. The mean is within $10\%$ of its minimum for every $\varepsilon$ between $10^{-3}$ and $10^{-2}$ in all four configurations, with the minimum between $0.003$ and $0.03$: the larger buffer of a tight budget and the redraws of a loose one cost about the same there. Outside that range both ends lose, by up to $16\%$ at $\varepsilon=10^{-4}$, where the buffer at $N=2^{24}$, $k=2^{8}$ is $34\%$ larger than at $10^{-3}$, and by $15$ to $30\%$ at $\varepsilon=0.3$, where a call redraws $0.2$ to $0.4$ times on average. The tail separates the two ends: for $\varepsilon\le10^{-2}$ the $99$th percentile stays within the run-to-run jitter of this GPU, $1.3$ to $2\times$ the median, whereas at $\varepsilon=0.1$ one call in ten runs the pass twice and the $99$th percentile rises to $2.2$ to $3.6\times$ the median, and to $3$ to $4.3\times$ at $\varepsilon=0.3$. A budget of $10^{-3}$ therefore costs at most $10\%$ in expectation against the fastest budget and keeps the latency of every call predictable, which is why we recommend it as a reasonable default.